# Artificial Intelligence-Enabled Intelligent Assistant for Personalized and Adaptive Learning in Higher Education

Ramteja Sajja[1,3], Yusuf Sermet[3], Muhammed Cikmaz[3], David Cwiertny[2,3,4,5], Ibrahim Demir[1,2,3]

[1] Department of Electrical Computer Engineering, University of Iowa
[2] Department of Civil and Environmental Engineering, University of Iowa
[3] IIHR Hydroscience and Engineering, University of Iowa
[4] Department of Chemistry, University of Iowa
[5] Center for Health Effects of Environmental Contamination, University of Iowa

**Abstract**
This paper presents a novel framework, Artificial Intelligence-Enabled Intelligent Assistant (AIIA), for personalized and adaptive learning in higher education. The AIIA system leverages advanced AI and Natural Language Processing (NLP) techniques to create an interactive and engaging learning platform. This platform is engineered to reduce cognitive load on learners by providing easy access to information, facilitating knowledge assessment, and delivering personalized learning support tailored to individual needs and learning styles. The AIIA's capabilities include understanding and responding to student inquiries, generating quizzes and flashcards, and offering personalized learning pathways. The research findings have the potential to significantly impact the design, implementation, and evaluation of AI-enabled Virtual Teaching Assistants (VTAs) in higher education, informing the development of innovative educational tools that can enhance student learning outcomes, engagement, and satisfaction. The paper presents the methodology, system architecture, intelligent services, and integration with Learning Management Systems (LMSs) while discussing the challenges, limitations, and future directions for the development of AI-enabled intelligent assistants in education.

## 1. Introduction

The landscape of higher education is experiencing a significant transformation, propelled by rapid advancements in digital technology and the evolving needs of a diverse and globally distributed student population (Altbach et al., 2009). Traditional teaching methods, while effective in many contexts, often struggle to provide personalized support and instant feedback, particularly in fields that demand a significant amount of text-based learning, critical thinking, and analytical skills (Means et al., 2009). These fields, such as Creativity and Critical Analysis, and Society and Culture, can pose challenges for students to master without adequate support (Holmes et al., 2019). This has led to a growing interest in exploring innovative solutions that can enhance the learning experience and outcomes for students in these fields, and beyond (Popenici and Kerr, 2017).

Artificial Intelligence (AI) and Natural Language Processing (NLP) have emerged as promising technologies with the potential to revolutionize the educational landscape. NLP and knowledge generation systems have been used actively for communicating data and information (Baydaroglu et al., 2023) in environmental (Sermet & Demir, 2018) and health (Zhang et al., 2023; Sermet & Demir, 2021) domains. The advent of AI-enabled tools, such as virtual teaching assistants (VTAs), offers a unique opportunity to bridge the gap between traditional teaching practices and the evolving needs of students (Winkler & Söllner, 2018). VTAs can provide personalized support, instant feedback, and adaptive learning experiences, thereby enhancing student engagement, satisfaction, and learning outcomes (Fryer et al., 2017).

Moreover, these AI-enabled solutions are not limited to text-based materials. Advanced deep learning models have been successfully used for synthetic image generation (Gautam et al., 2022), image data augmentation (Demiray et al., 2021) and image analysis (Li & Demir, 2023). They can also support learning in areas such as coding, mathematics and statistics, and even visual inputs. By leveraging AI and NLP, VTAs can interpret and provide feedback on code snippets, mathematical equations, and statistical models. They can also process and respond to visual inputs such as diagrams, charts, images, videos, and maps, further expanding their utility in diverse learning contexts.

Web technologies play a crucial role in embedding Large Language Models (LLMs) and chatbots into the intricate fabric of modern engineering education, catering to a myriad of specialized domains. In the realm of advanced modeling (Ewing et al., 2022) and analysis tools (Sit et al., 2021), web platforms enable real-time processing and intuitive visualization of complex engineering problems, enhancing students' ability to grasp and manipulate sophisticated models. When diving into the vast sea of programming libraries, as documented by Ramirez et al. (2022; 2023), web technologies make it feasible to offer on-the-spot guidance, code suggestions, and troubleshooting advice, assisting budding engineers in seamlessly navigating and utilizing these libraries.

Furthermore, the convergence of LLMs, chatbots, and web platforms has been instrumental in redefining pedagogical methods. Here, web-hosted chatbots, powered by LLMs, can simulate ethical dilemmas, guide reflections, and provide instant feedback, ensuring that future engineers not only excel in their technical prowess but also uphold the ethical standards of their profession. However, the effectiveness of VTAs in supporting students' learning needs in these diverse fields, where multi-modal data plays a significant role, remains an area ripe for exploration. The potential of VTAs to enhance learning outcomes across a wide range of disciplines and learning formats underscores the need for further research and development in this growing field.

This study introduces a novel web-based framework for an AI-enabled Virtual Teaching Assistant (AIIA), designed to enhance student learning in qualitative disciplines. The AIIA, built with a NodeJS backend, leverages the power of AI and Natural Language Processing (NLP) to create an interactive and engaging platform. This platform is engineered to reduce the cognitive load on learners by providing easy access to information and facilitating knowledge assessment. The AIIA's capabilities include understanding and responding to student inquiries, generating quizzes and flashcards, and delivering personalized learning support tailored to individual needs and learning styles. By presenting this innovative framework, this paper contributes to the ongoing efforts to integrate AI-enabled technologies and web systems into education, aiming to improve the effectiveness of learning support in qualitative fields.

The potential impact of this research is significant, as it can provide valuable insights into the design, implementation, and evaluation of AI-enabled VTAs in higher education. The findings of this study can inform the development of innovative educational tools that can enhance student learning outcomes, engagement, and satisfaction. Furthermore, the research can contribute to the broader discourse on the integration of AI and NLP in education, providing empirical evidence on the effectiveness of these technologies in enhancing teaching and learning practices.

The remainder of this article is organized as follows. Section 2 summarizes the relevant literature and identifies the knowledge gap. Section 3 presents the methodology of the design choices, development and implementation of a course-oriented intelligent assistance system. Section 4 describes the features implemented for both the instructors and students. Section 5 discusses the strengths, limitations, and future directions. Section 6 concludes the articles with a summary of contributions.

## 2. Related Work

The literature on the application of AI in education has grown substantially in recent years, reflecting the increasing interest in this field. In this section, we systematically review existing literature, specifically focusing on the use of Virtual Teaching Assistants (VTAs) in higher education and natural language communication and identify the knowledge gap that justifies the present research.

A critical paper by Huang, Saleh, and Liu (2021) provides an overview of AI applications in education, including adaptive learning, teaching evaluation, and virtual classrooms. This study highlights the potential of AI to promote education reform and enhance teaching and learning in various educational contexts. Essel et al. (2022) presents a study on the effectiveness of a chatbot as a virtual teaching assistant in higher education in Ghana, demonstrating that students who interacted with the chatbot performed better academically compared to those who interacted with the course instructor. This empirical evidence supports the potential of VTAs to improve student academic performance. Crompton and Song (2021) provide a comprehensive overview of AI in higher education, discussing its potential in various aspects such as bespoke learning, intelligent tutoring systems, facilitating collaboration, and automated grading. This paper contributes to the broader discourse on the integration of AI and natural language processing in education.

In addition to these empirical studies, several recent publications delve further into AI-enhanced educational systems. Akgun and Greenhow (2021) discuss the ethical challenges of using AI in education and the potential applications, such as personalized learning platforms and automated assessment systems. Ewing and Demir (2021) discuss ethical challenges in engineering decision making using AI from educational perspective. Bahja (2020) offers a comprehensive explanation of Natural Language Processing (NLP), its history, development, and application in various industrial sectors. In the context of large language models, Neumann et al.'s (2023) paper explores the potential approaches for integrating ChatGPT into higher education, focusing on the effects of ChatGPT on higher education in software engineering and scientific writing. Pursnani et al. (2023) assessed the performance of ChatGPT on the US fundamentals of engineering exam (FE Exam) and did a comprehensive assessment of proficiency and potential implications for professional environmental engineering practice. Sajja et al. introduce an AI-augmented intelligent educational assistance framework based on GPT-3 and focused on curriculum- and syllabus-oriented support, which automatically generates course-specific intelligent assistants regardless of discipline or academic level.

Furthermore, Tack and Piech (2022) examine the pedagogical abilities of Blender and GPT-3 in educational dialogues, finding that conversational agents perform well on conversational uptake but are quantifiably worse than real teachers on several pedagogical dimensions, especially helpfulness. Lee (2022) explores the potential of ChatGPT in medical education, discussing its potential to increase student engagement and enhance learning, as well as the need for further research to confirm these claims and address the ethical issues and potential harmful effects. Perkins et al. (2022) examine the academic integrity considerations of students' use of AI tools using large language models, such as ChatGPT, in formal assessments, emphasizing the need for updated academic integrity policies to consider the use of these tools in future educational environments. Lastly, Audras et al. (2021) discuss the potential application of VTAs to reduce the burden on teachers across secondary schools in China, emphasizing the need for careful design and attention to student support.

In conclusion, the existing literature highlights the potential benefits and challenges of using AI-based VTAs in higher education. While there is a growing body of research on the design, implementation, and effectiveness of VTAs, several key areas remain to be addressed in the literature. These include the scalability and adaptability of such systems across diverse learning contexts, their potential impact on the future trajectory of higher education, and the integration of these systems with Learning Management Systems (LMS).

Furthermore, most studies have not considered the incorporation of class recordings and class interactions in their AI-based solutions, which could potentially enrich the knowledge base of VTAs and provide a more comprehensive learning experience for students. Additionally, existing literature has not extensively addressed the need for a solution that caters to both students and instructors, striking a balance between personalized assistance and instructor support.

Another critical aspect that has not been adequately addressed in the literature is the potential for cheating and academic dishonesty that may arise with the use of AI-based VTAs. Ensuring academic integrity and preventing cheating should be an integral part of any AI-enabled educational solution, yet there is a dearth of research exploring effective prevention mechanisms (Kasneci et al., 2023).

The current study aims to address these gaps by designing, implementing, and evaluating an AI-enabled Intelligent Assistant (AIIA) for personalized and adaptive learning in higher education. Proposed AIIA seeks to seamlessly integrate with existing LMS, utilize class recordings and class interactions, cater to the needs of both students and instructors, and incorporate measures to ensure academic integrity and prevent cheating. By addressing these knowledge gaps, this study contributes to the ongoing efforts towards the development and implementation of effective AI-based educational solutions in higher education.

## 3. Methodology

The primary objective of this research is to address the growing need for innovative educational solutions in higher education, catering to the diverse needs of learners and fostering an inclusive, equitable, and engaging learning environment. By harnessing the power of conversational AI and advanced natural language processing techniques, the proposed framework seeks to improve learning experiences and outcomes in postsecondary education, while bridging learning gaps and facilitating continuous learning through flexible educational pathways. The AIIA aims to be discipline-independent, scalable, and seamlessly integrated across institutions, thereby unlocking its potential to impact a broad spectrum of students and educators.

The transformative nature of AIIA lies in its convergence of advanced AI technologies with effective educational principles, promoting self-regulated learning, fostering student-faculty communication, encouraging collaboration, and enhancing access to learning resources. VirtualTA system offers a range of benefits for students and higher education, including:

a) *Enhanced Learning Experience*: Providing a personalized and interactive learning experience, where students can ask questions, seek clarifications, and access relevant resources in real-time.
b) *Instant Access to Information*: Enabling efficient knowledge acquisition by quickly retrieving information from various course resources.
c) *On-Demand Support*: Offering 24/7 assistance, promoting self-directed learning, and empowering students to take ownership of their education.
d) *Consistency and Accuracy*: Delivering reliable information, reducing the risk of incorrect or conflicting answers.
e) *Adaptive Learning*: Facilitating personalized learning paths, catering to diverse needs, and promoting effective knowledge retention.
f) *Multilingual Support*: Expanding the AIIA's capabilities to include support for multiple languages, ensuring that students from diverse linguistic backgrounds can effectively engage with and benefit from the AI-enabled assistant.
g) *Expansion of Access*: Integrating into digital platforms for broader access to quality education and enabling remote learning for students worldwide.
h) *Automation of Administrative Tasks*: Freeing up instructors' time for higher-value activities, such as facilitating discussions and providing personalized guidance to students.

i) *Personalized Learning, Continuous Assessment and Feedback*: Utilizing adaptive self-learning mechanisms and providing timely and constructive guidance for students to take an active role in their learning journey.
j) *Addressing Emotional and Social Aspects of Learning*: Incorporating emotional intelligence and social awareness into the AIIA, enabling it to recognize and respond to students' emotional states and provide empathetic support.

By incorporating a range of AI-enabled functionalities, AIIA seeks to harness the "Protégé Effect", ultimately contributing to increased learning proficiency and mitigating educational inequality. Additionally, the integration of AIIA into various communication channels ensures accessibility for students of diverse backgrounds, further promoting equity in higher education. This research is poised to make a significant contribution to the ongoing discourse on the integration of AI and natural language processing in education, shaping the future trajectory of higher education and empowering the next generation of professionals.

## 3.1. Natural Language Inference

Large language models (LLMs) use deep learning algorithms to analyze and generate human language, having applications ranging from chatbots to translation systems. Trained on extensive text data, LLMs like GPT-3 (Brown et al., 2020), GPT-2 (Radford et al., 2019), PaLM (Chowdhery et al., 2022), BERT (Devlin et al., 2019), XLNet (Yang et al., 2020), RoBERTa (Liu et al., 2019), ALBERT (Lan et al., 2020), and T5 (Raffel et al., 2020) can generate responses emulating human-like communication.

OpenAI's Generative Pretrained Transformer 3.5 (GPT-3.5) serves as a leading-edge autoregressive language model, capable of synthesizing textual content akin to human composition. The model's versatility is demonstrated through its adaptability to an array of applications, a feature attributable to few-shot learning and fine-tuning methods. Few-shot learning allows the model to tackle unfamiliar tasks with minimal example provision, leveraging its extensive pre-training on a wide variety of internet text data. Conversely, fine-tuning involves training the model on a significant number of task-specific examples, thereby augmenting its performance in distinct application domains and obviating the need for examples in the prompt.

For the study, we selected GPT-3.5 due to its user-friendly API and advanced natural language processing capabilities. We utilized the "text-davinci-003" variant, a GPT-3 model built on InstructGPT (Ouyang et al., 2022), appreciated for its few-shot learning and fine-tuning capabilities. Additionally, we also employed other GPT-3.5 models, including "gpt-3.5-turbo" for text completions, a fine-tuned Davinci model for query classification, and a Fine-Tuned Curie model for open-ended question generation.

### 3.1.1. Text Embeddings

In the field of Natural Language Processing (NLP), embeddings are numerical representations that help computers understand the meaning and context of different concepts. They are used in various applications such as search functions, recommendations, and categorizations, offering significant benefits. For this research, we used OpenAI's text-embedding-ada-002 (Greene et al., 2022) model to convert various classroom materials - including assignments, announcements, lecture notes, forum posts, and recordings - into text embeddings. This model is well-suited for dealing with long documents and provides embeddings with 1,536 dimensions (Greene et al., 2022). This conversion process forms the basis for the development of a search algorithm that uses cosine similarity to find documents most relevant to a user's query. By turning course materials into embeddings, we can efficiently find and retrieve relevant information without the need to manually search each document. Our approach highlights the effectiveness of using embeddings in NLP tasks, particularly in the context of document search and retrieval. This contributes to a system capable of providing accurate and specific responses.

In assessing the semantic similarity between two vectors, it is essential to compare the generated word embeddings. Cosine similarity, a measure calculating the cosine of the angle between two vectors, is often employed for this purpose. This process essentially conducts a dot product operation between the vectors. When the vectors perfectly align at 0 degrees, the cosine value becomes 1, representing complete similarity. For angles other than 0 degrees, the cosine value drops below 1, further decreasing as the angle widens. Thus, the larger the cosine similarity, the more closely aligned or "similar" the two-word embeddings are (Gunawan et al., 2018).

This similarity metric underpins the search algorithm's operation, aiding in identifying documents most relevant to a user's query. The algorithm prioritizes vectors with higher cosine similarity values, thereby enhancing search precision. By employing cosine similarity, the system identifies the most appropriate match within the course data embeddings and curates a list of 10 documents with the highest correlation to the user's query. Only those exhibiting a similarity score above 75% are retained, with their text forming the context for responding to the user's query.

To uphold a high accuracy level prompt engineering techniques are applied to prevent hallucination, ensuring VirtualTA does not furnish incorrect responses. In scenarios where the model lacks confidence in its response, it refrains from providing an answer and instead conveys a message of uncertainty. This mechanism plays a crucial role in averting the propagation of erroneous information, thereby ensuring that users receive only accurate and reliable information.

### 3.1.2. Transcription and Speaker Diarisation

In response to the widespread shift to remote learning following the Covid-19 pandemic, the recording of lectures and classes has become a prevalent practice. Recognizing the potential of these rich, yet underutilized resources, our research aimed to incorporate these recorded materials into VirtualTA system via Automatic Speech Recognition (ASR) technology. To this end, we adopted Whisper, an ASR system developed by OpenAI (Radford et al., 2022), trained on a substantial corpus of multilingual and multitask supervised data, totaling 680,000 hours.

Whisper allows us to transcribe speech from recorded classes into textual data, which can subsequently be processed and analyzed by the system.

This integration significantly expands the data pool available for analysis and query resolution, enhancing our ability to support student learning. By including speech data from class recordings, we not only augment the comprehensiveness of our responses, but also facilitate a more effective transfer of knowledge. This approach underscores our commitment to fully exploiting available resources, as we continuously strive to enhance the learning experiences of students.

### 3.2. System Architecture

The System Architecture of the Artificial Intelligence-Enabled Intelligent Assistant (AIIA) (Figure 1) framework serves as the foundation for its operation and functionalities within the higher education context. This architecture comprises four primary components: 1) Data Retrieval, which focuses on obtaining and processing various data resources through CANVAS integration and transcription services; 2) Core Framework, which encompasses the design and implementation of language services, system design, and server management to ensure efficient operation; 3) Intelligent Services, which includes the Virtual TA, Study Partner, and Instructor Assistant functionalities that cater to the diverse needs of students and instructors; and 4) Communication, which facilitates seamless interaction between the system and its users through web-based chatbots, accessibility features, and multi-platform support. This comprehensive architecture enables the AIIA framework to deliver personalized and adaptive learning experiences, fostering enhanced engagement and improved learning outcomes in higher education environments.

### 3.2.1. Data Resources: Categorization, Prioritization, and Transparency

VirtualTA system utilizes a range of data resources in its operation, primarily targeting elements intrinsic to the course structure. Table 1 provides an overview of these key resources, which include but are not limited to Assignments, Announcements, Discussions, Lectures, and External Reading Materials. Each resource type plays a distinct role within VirtualTA's architecture, contributing to the system's ability to accurately respond to student queries. For example, Assignments are used to gauge the context of the student's query, while Announcements ensure that the system can provide the most up-to-date information regarding the course. To manage the various resource types efficiently, a unique data structure has been implemented. This structure not only categorizes the resources but also prioritizes them based on their relevance to the learning objectives of the course.

For instance, primary resources such as Lectures are given a higher priority compared to secondary resources like External Reading Materials. This hierarchical approach ensures that the system first seeks answers from the most critical resources, thereby enhancing the accuracy and reliability of the responses generated. Furthermore, this structured approach also offers traceability, allowing the system to identify and disclose the resources it used to derive an answer. This feature adds a layer of transparency to the system's operation, providing users with insights into the sources of the information supplied, and contributing to their confidence in the system's responses.

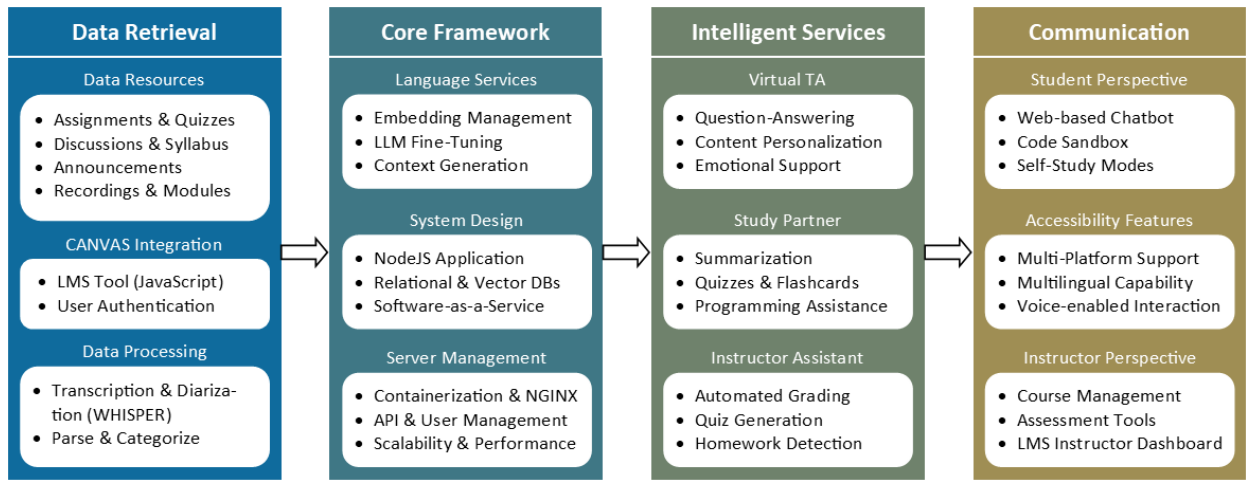

Figure 1. System architecture of VirtualTA

Table 1: Overview of Key Course Resources Used by VirtualTA System

| Resource | Description |
| --- | --- |
| Assignments | *Instructor-issued tasks intended for gauging students' comprehension and course progression. VirtualTA system utilizes this data not only for tracking assignment deadlines but also for understanding the context of the assignment. This assists in detecting whether a student's query is assignment-related, enabling more targeted and effective assistance.* |
| Discussions | *Forums for students to engage in discourse about course-related topics. Utilized by VirtualTA to answer questions and provide insights.* |
| Announcements | *Vital notifications issued by instructors regarding course alterations, deadlines, or events. VirtualTA system maintains an updated record of these announcements, facilitating accurate and timely responses to students' inquiries with the most current information available.* |
| Lectures | *Recorded or live teaching sessions that deliver course content. VirtualTA uses lecture transcripts to answer queries related to course content.* |
| Reading Materials | *Required or recommended readings for the course. VirtualTA can use these materials to answer relevant questions, summarize complex readings, or create reading plans.* |
| Quizzes | *Short assessments designed to test a student's grasp of recent course material. VirtualTA can assist students in quiz preparation by generating similar questions for practice.* |

### 3.2.2. Knowledge Base Generation

VirtualTA chatbot is empowered by a comprehensive knowledge generation process that incorporates extraction, parsing, and encoding of resources obtained from a learning management system (LMS). This extensive process comprises a series of steps that ensure optimal utilization of available resources and enhance the efficiency of the chatbot. The initial stage revolves around acquiring a wealth of information encapsulated in documents such as lecture files, lecture recordings, and reading materials. The acquisition process also includes additional resources such as assignments, discussion board entries, quiz information, and course announcements.

Upon data acquisition, a parsing technique is employed where applicable. This technique, primarily applied to file-format data, meticulously breaks down the text into manageable chunks, each consisting of approximately 800 characters. This operation is conducted with the utmost care to ensure words and sentences remain intact, thus enabling the production of coherent and meaningful blocks of information. However, it is important to note that some resources bypass the parsing stage. Announcements and assignment details, typically supplied by the LMS API, are already presented in a structured JSON format. Their inherent structure eliminates the need for parsing, streamlining the process and enhancing efficiency.

Subsequent to data extraction and parsing, the resulting information is encoded into text embeddings. In this transformative phase, the 800-character blocks are metamorphosed into high-dimensional vector representations. This process not only preserves but also enhances the semantic richness of the course content. The set of embeddings created forms the fundamental knowledge base of VirtualTA chatbot. These embeddings enable the chatbot to offer advanced, intelligent services to both students and instructors as detailed in Section 4.

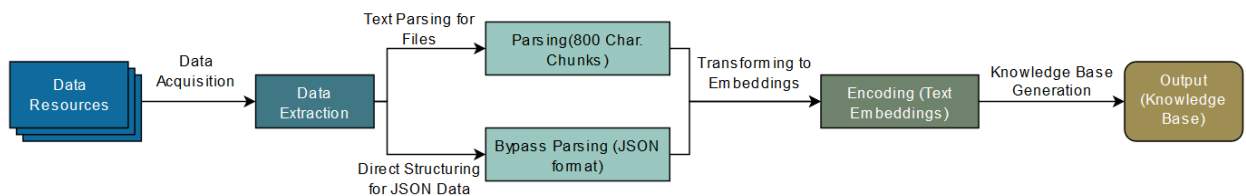

Figure 2: Autonomous knowledge base population

Figure 2 offers a visual representation of the autonomous knowledge base generation process. The sequence begins with the acquisition of data resources (documents and other elements), proceeds through data extraction and parsing (or bypassing parsing), followed by the generation of embeddings. This culminates in the formation of the output – a robust, dynamic knowledge base.

### 3.2.3. Advanced Query Interpretation and Response Generation

In the core application of the system, a multistage process is deployed prior to generating a response to a user's query, a process integral to the efficient functioning of the system. This process is elucidated in the subsections below.

Query Classification: The system begins by distinguishing the nature of the user's query, a process known as query classification. This phase discerns the type and intent of the question, facilitating a more focused and relevant response.

Context Generation and Embedding Matching: Subsequent to classification, the system transitions to the context generation phase. The user's question is transformed into a text embedding, a vectorized representation that allows the query to be accurately compared to the existing knowledge base. The system employs cosine similarity to identify the closest match within the course data embeddings. A list of the ten documents with the highest correlation to the user's query is curated, retaining only those with a similarity score exceeding 75%. The text from these documents is then utilized to form the context for the system's response.

Response Generation and Hallucination Mitigation: To uphold the accuracy and relevance of the system's output, we selectively apply fine-tuning models to certain features. These models assist in generating various question types, including open-ended, true/false, and multiple-choice questions. In the subsequent stage, prompt engineering techniques are utilized to mitigate hallucination — the generation of incorrect or irrelevant information — thereby ensuring the Virtual TA does not produce inaccurate responses.

Error Prevention Mechanism: A unique feature of the system is its built-in error prevention mechanism. If the model lacks confidence in the accuracy of its response, it refrains from providing an answer. Instead, it communicates a message such as "I'm not sure," which aids in preventing the dissemination of erroneous information. This feature ensures that users only receive information that is both accurate and reliable.

User Intent Fulfillment: This is the final stage, where the classified query (from the Query Classification step) is executed, fulfilling the user's specific intent. For instance, if the user wants a question answered, a topic summarized, automatic code generation, question generation, or an essay outline on a given topic, the system will proceed accordingly to meet the user's needs.

### 3.2.4. Cyberinfrastructure and Integration

The proposed framework is grounded on a centralized, web-based cyberinfrastructure responsible for various tasks, including data acquisition, training of deep learning models, storage and processing of course-specific information, and hosting the generated chatbots for utilization in a frontend application. The cyberinfrastructure comprises an NGINX web server and NodeJS-based backend logic, bolstered by a PostgreSQL database, caching mechanisms, and modules for user and course management. The heart of this setup is the intelligent assistant, architected on a Service-Oriented Architecture (SOA) that enables plug-and-play integration with any web platform supporting webhooks. The key elements of this section include a student chat interface with multimodal responses, an instructor interface for resource management and analytics, a new JS library for LMS integration, and the Whisper-based Speech API for transcription services.

*Student Chat Interface*: The AIIA system features a web-based chat interface with multimodal responses, allowing for efficient communication between students and VirtualTA.

Interaction with VirtualTA system is enabled through a specially developed API, which retrieves the system's responses for presentation to the user via the chatbot. This chat interface is integrated directly into Canvas, providing students with easy access to the AI assistant and encouraging them to engage with the LMS more frequently. By embedding the chatbot within the familiar Canvas platform, the AIIA system ensures a smooth and seamless user experience for students.

*Instructor Dashboard*: The administrative interface, built using React, empowers instructors to manage the resources utilized by the AIIA system. This interface allows instructors to enable or disable specific resources, providing control over the information accessible to students. Additionally, the instructor dashboard offers access to analytics, enabling instructors to monitor student engagement and performance.

*LMS Integration*: To facilitate seamless integration with various LMSs, particularly with Canvas, a new JavaScript library has been developed. This library enables the AIIA system to interact with multiple courses, retrieve and preprocess relevant data, and regenerate course embeddings as needed. By providing compatibility with a broadly adopted LMS, the AIIA system ensures its adaptability and applicability across diverse educational settings.

*Speech API*: The Speech API, implemented as a backend-service in Python and served via Flask, is based on WHISPER and pyannote (i.e., a Python package for neural speaker diarisation) and plays a pivotal role in providing transcription services for the AIIA system. The API offers a variety of tailored endpoints, catering to different transcription use cases including transcribing video content from Canvas file URLs, with or without timestamps, and transcribing videos from YouTube URLs or other specified URLs, also with or without timestamps. These versatile endpoints enable the AIIA system to efficiently transcribe video content from a diverse range of sources, ensuring that the AI assistant has access to a comprehensive array of course materials and information to provide accurate, context-aware responses to student queries.

## 4. Results

In this section, we present the intelligent services and enhancements implemented in the system to cater to the needs of both students and instructors. These advancements aim to augment the learning experience by providing students with valuable tools and resources, while also assisting instructors in their instructional tasks and assessments. The student-oriented enhancements encompass features such as Dynamic Flashcard Integration, Automated Assessment: Intelligent Quiz Generation and Auto-grading, Automated Question-Answering on Course-Related Topics, Embedded Sandbox Integration within the Chatbot Interface, summarization of course content, and context-aware conversation. On the other hand, the instructor-focused enhancements include an Auto-Evaluator for Streamlined Assignment Assessment, Automated Homework Detection Mechanism to promote independent learning, and Automated Generation of Diverse Assessment Questions. By incorporating these intelligent services, the system aims to create a dynamic and interactive learning environment, supporting both students and instructors in their academic pursuits.

## 4.1. Student-Oriented Enhancements

This section highlights the student-oriented enhancements integrated into the system to enhance the learning experience and support students in their academic pursuits. These enhancements encompass dynamic flashcard integration, automated assessment with intelligent quiz generation and auto-grading, automated question-answering on course-related topics, embedded sandbox integration within the chatbot interface, summarization of course-related topics, and context-aware conversation. By incorporating these features, the system aims to provide students with valuable study resources, efficient assessment tools, prompt information retrieval, programming assistance, condensed topic summaries, and personalized communication. These enhancements contribute to creating an engaging and effective learning environment that fosters comprehension, active participation, and self-assessment for students.

### 4.1.1. Dynamic Flashcard Integration

A notable addition to the system is the incorporation of a flashcard feature, enabling students to request flashcards on any topic within the course to support their preparation. This feature closely resembles traditional flashcards, with the front side presenting a question and the flip side revealing the answer. In the implementation, the flashcards encompass both true/false questions and open-ended questions. Furthermore, each answer is accompanied by a detailed explanation or reasoning, elucidating the rationale behind the given answer. This feature serves to enhance students' understanding and retention of course concepts, providing them with a valuable study resource. The flashcards depicted in Figure 3 present a format wherein the front side contains a question, while the flip side reveals the corresponding answer along with the underlying reasoning.

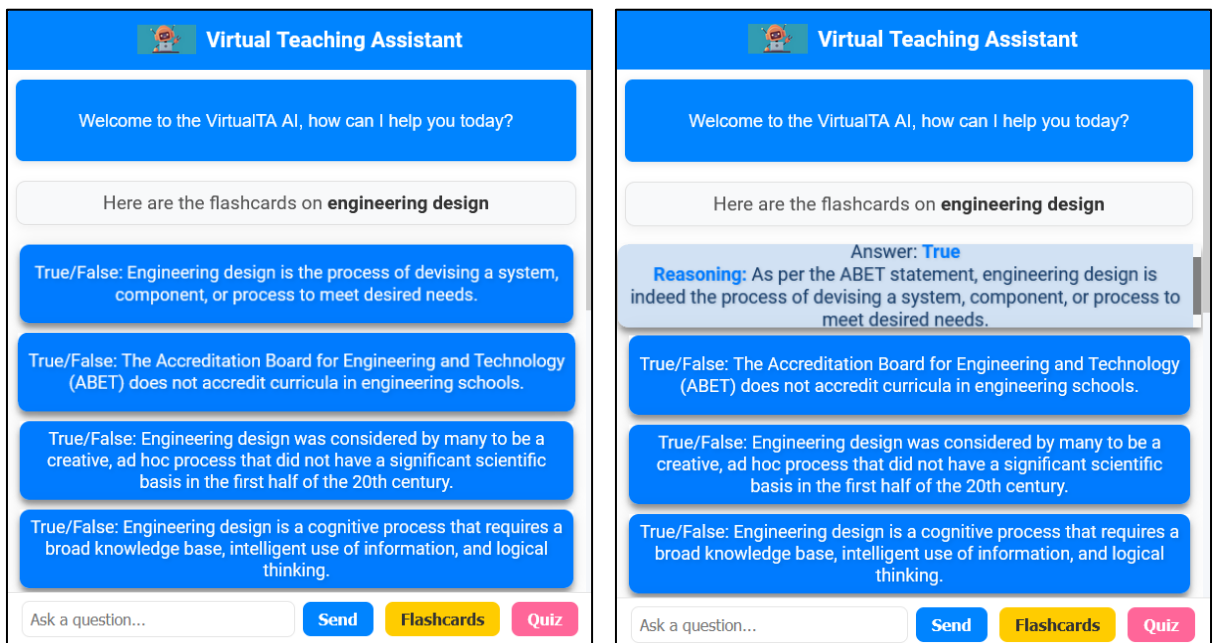

Figure 3: Flashcard Functionality of VirtualTA

### 4.1.2. Automated Assessment: Intelligent Quiz Generation and Auto-grading

In addition to the flashcard functionality, the system also includes a quiz feature that allows users to request quizzes on specific topics from the course. This feature enables students to test their knowledge and understanding of the subject matter. The quizzes consist of both true/false questions and open-ended questions, providing a comprehensive assessment of the students' grasp of the material. To enhance the user experience, we have implemented an auto-grading system for the quizzes. Once the student submits their answers, the system automatically evaluates their responses. The system provides immediate feedback by indicating whether each answer is correct or incorrect. In cases where the answer is incorrect, an explanation is provided to help the student understand the correct response and the underlying reasoning.

By incorporating this quiz feature with auto-grading functionality, we aim to foster an interactive learning experience that promotes active participation and self-assessment. Students can gauge their progress, identify areas of improvement, and reinforce their understanding through the provided explanations. Figure 4, displayed below, showcases the quiz functionality, also referred to as the self-assessment functionality. In this feature, users are presented with a question and have the ability to input their answer. Upon clicking the submit button, the system evaluates the correctness of the response and provides accompanying reasoning or explanations.

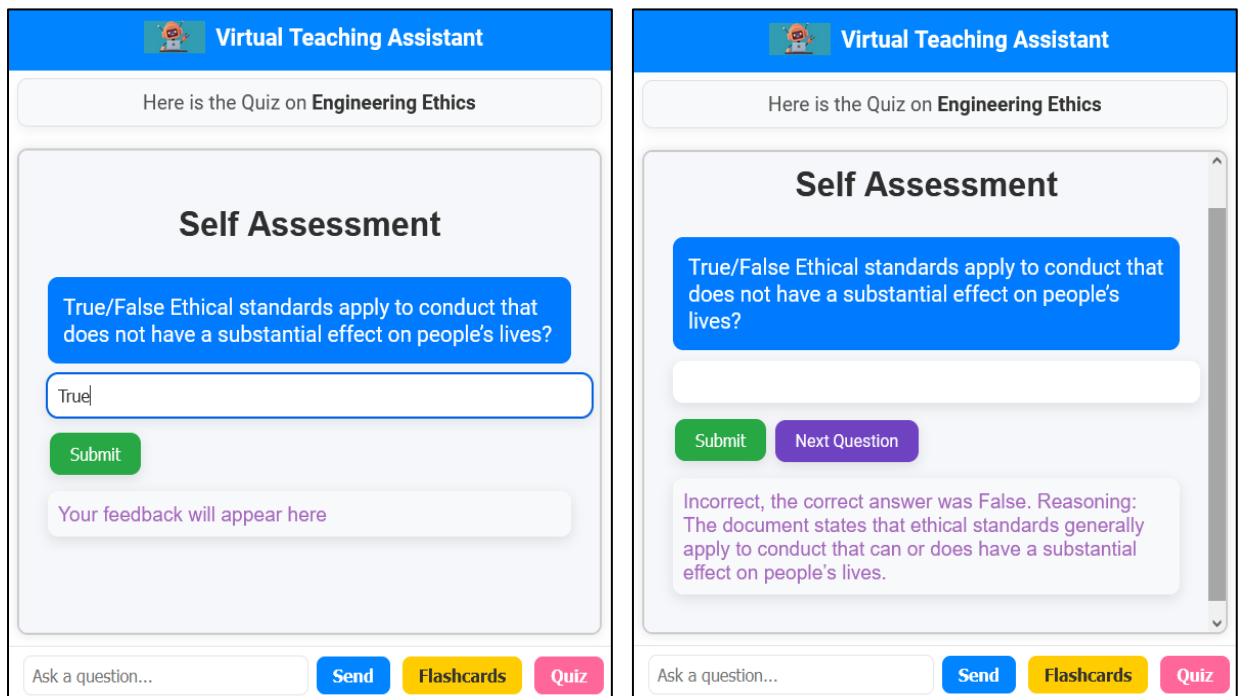

Figure 4: Quiz Generation and Auto-Grading Functionality of VirtualTA

### 4.1.3. Automated Question-Answering on Course-Related Topics

The system incorporates a feature that enables students to ask questions pertaining to administrative or course content topics. This functionality is specifically designed to streamline the process of obtaining answers to common inquiries, thereby enhancing the overall learning experience for students. By leveraging available information within the system's knowledge base, automated response mechanism ensures prompt and accurate responses to a wide range of queries. Students can seek information on various administrative aspects, such as important dates or course logistics, as well as delve into specific course content topics, seeking clarification or further insights. For instance, a student might ask about upcoming midterm dates or inquire about the engineering design process. When the necessary information is present within the system's knowledge base, the system generates automated responses that directly address the student's query, providing the relevant details or explanations.

This feature not only expedites the process of obtaining information but also empowers students to take charge of their learning journey. By leveraging automation and readily available knowledge, the system offers students a convenient and efficient means of accessing accurate responses to their questions, thereby fostering an enhanced learning experience. Figure 5, depicted below, exemplifies the question-answering feature. When a user poses a question, the system promptly responds with an answer, accompanied by a disclaimer. This disclaimer includes the confidence percentage of the response and provides information regarding the source from which the information was obtained. In cases where the answer is automatically generated, indicating a lack of matching documents, the system acknowledges that it attempted to answer the question autonomously and advises users to consult with an expert for matters of significant importance.

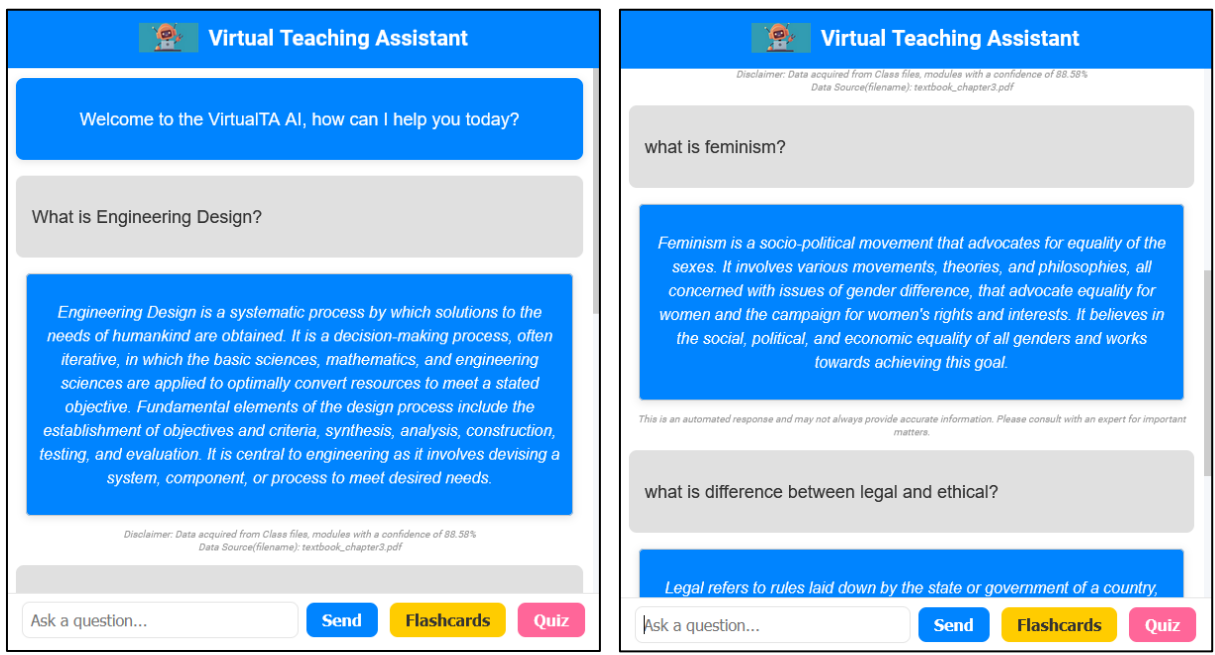

Figure 5: Web based chatbot user interface with questions and answers

### 4.1.4. Embedded Sandbox Integration within the Chatbot Interface

To enhance the learning experience and cater to students with varying levels of programming proficiency, we have integrated a coding sandbox environment directly into the chatbot interface. This feature allows users to seek assistance with programming-related queries and provides a convenient platform for code execution and clarification. Whether it is a programming course or a non-CS domain, students can ask for guidance or clarification on code snippets. The system automatically detects the programming language being used and, upon the user's request to "run code," opens up a coding sandbox environment within the chatbot itself. This eliminates the need for students to have prior knowledge of integrated development environments (IDEs) or programming tools.

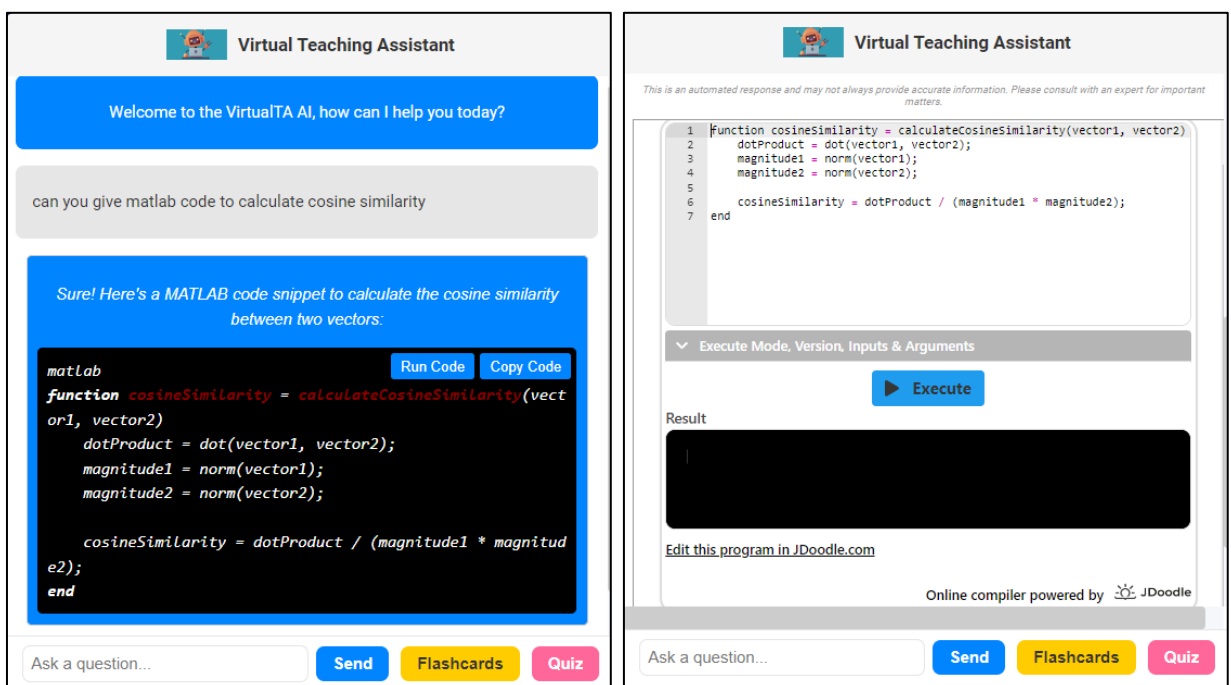

Figure 6: Code Sandbox Environment within VirtualTA

By offering this integrated coding sandbox, we aim to provide a user-friendly and accessible platform for students to experiment with and execute basic code related to their courses. This feature is particularly valuable for beginners or individuals unfamiliar with traditional coding environments, as it allows them to interact with code directly within the chatbot interface. It facilitates quick testing and understanding of programming concepts, promoting a more interactive and engaging learning experience. Figure 6, depicted above, showcases the seamless integration of a sandbox environment. It displays the MATLAB code corresponding to the user's query. Upon clicking the "Run Code" button, the code is automatically transported to an Integrated Development Environment (IDE) where users can conveniently execute the function directly within the environment.

### 4.1.5. Summarization of Course-Related Topics

To facilitate students' access to condensed information on specific topics of interest, we have implemented a summarization feature in the system. This feature allows students to request a summary of a particular topic, enabling them to quickly grasp the key points without having to go through extensive materials. Leveraging the power of GPT-3.5 models, the system is capable of understanding student queries pertaining to specific topics, such as feminism or engineering design. Upon receiving a query, the system utilizes its knowledge base to generate a relevant and concise summary that encapsulates the essential information related to the topic. This summary is then presented to the student, providing them with a quick and efficient way to obtain an overview of the desired subject matter.

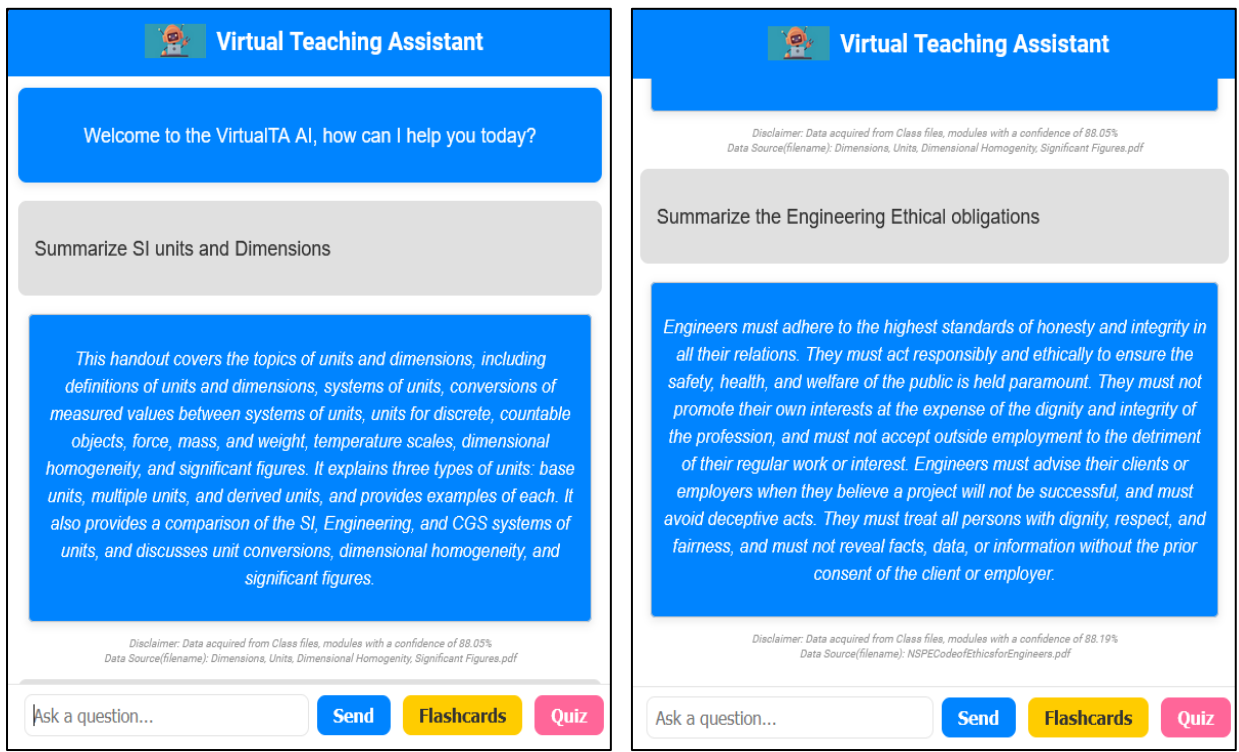

Figure 7: Summarization functionality of VirtualTA on Course Content

Figure 7, depicted above, exemplifies the summarization functionality. This feature enables the system to generate concise summaries of information related to any topic in the context of the course. When a user poses a question, the system leverages its summarization capabilities to provide a condensed summary as an answer. Similar to the question-answering feature, a disclaimer accompanies the response, indicating the confidence level and information source. It is important to note that the summarization feature aims to provide a brief overview, and for more comprehensive or critical matters, consulting an expert is recommended.

### 4.1.6. Context-Aware Conversation

To provide a more engaging and personalized experience, the system is designed to replicate the communication style of the student for better understanding and empathy. By analyzing the student's language patterns and preferences, the system adapts its responses to align with the student's communication style. To manage the conversation history effectively and ensure token limits are handled appropriately, we employ a dynamic rewriting and rotation approach to maintain relevant context while interacting with large language models (LLMs). Furthermore, the system is designed to adopt an uplifting, helpful, and empathetic persona. It aims to provide guidance, support, and relevant information to the student in a positive and constructive manner. In addition to adapting the communication style, VirtualTA system also incorporates techniques to identify the emotional state of the student. This capability enables the system to recognize when students are in need of empathy and understanding, allowing it to tailor its responses accordingly to provide the appropriate level of emotional support.

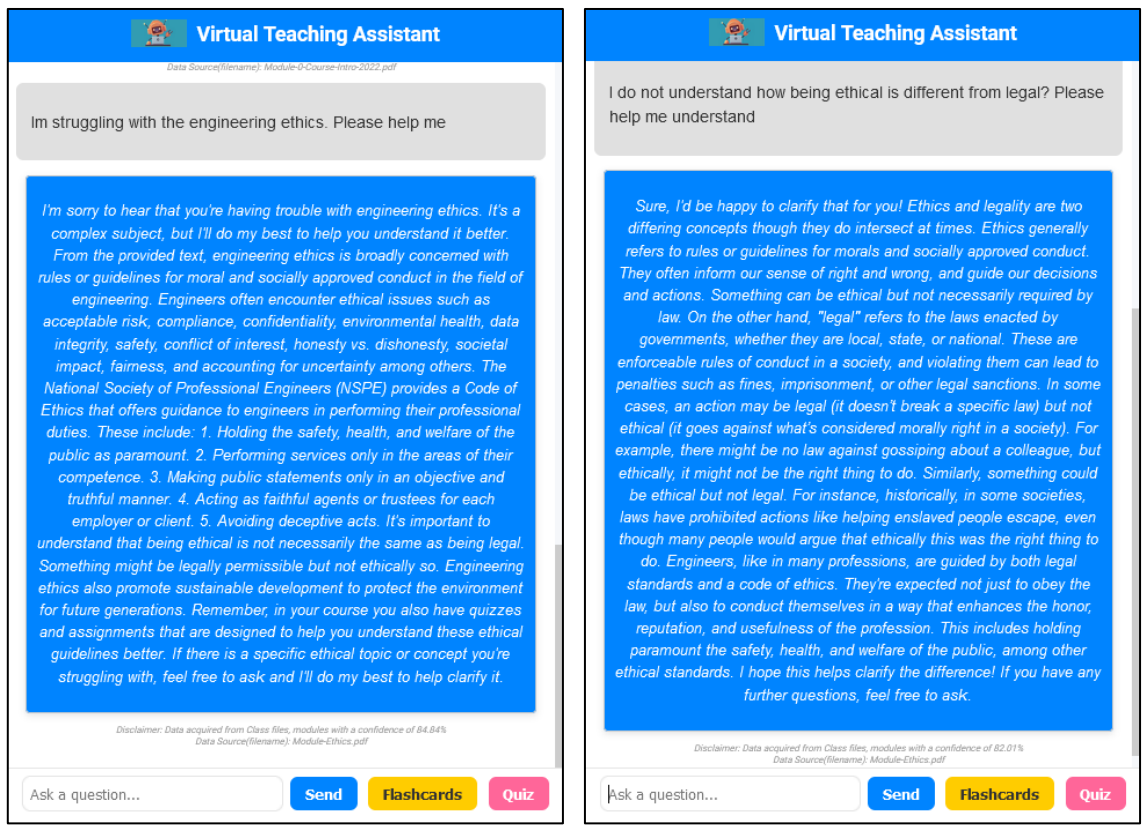

Figure 8: Context-aware replies by VirtualTA

In Figure 8, depicted above, highlights the context-aware capabilities of VirtualTA. This system possesses the ability to discern when a student is facing challenges or requires empathy and understanding. Instead of offering a straightforward response, the VirtualTA acknowledges the user's feelings, adopting an empathetic stance. Furthermore, it reinforces its continuous availability and commitment to assisting the student.

## 4.2. Instructor-Focused Enhancements

This section highlights the instructor-focused enhancements incorporated into the system to streamline instructional tasks and assessments. These enhancements include an auto-evaluator for efficient assignment assessment, an automated homework detection mechanism to promote independent learning and academic integrity, and the automated generation of diverse assessment questions. By integrating these features, the system aims to support instructors in their grading process, facilitate a comprehensive learning experience, and save time in question preparation for assessments.

### 4.2.1. Auto-Evaluator for Streamlined Assignment Assessment

In order to assist instructors in the grading process, we have incorporated an auto-evaluator feature for assignments. This feature allows instructors to provide us with the solutions, including the questions and their corresponding correct answers. Additionally, instructors can submit the students' responses to the questions, with the ability to upload PDF files for convenience. The system then automatically evaluates the submitted answers against the provided solutions, providing scores for each question. Furthermore, the system generates reasoning and explanations to justify the assigned scores, aiding instructors in understanding the evaluation outcomes. It is important to note that the auto-evaluator is not intended to replace the instructor's grading but rather to provide valuable insights and facilitate decision-making during the grading process. Ultimately, it aims to assist instructors by providing an initial evaluation and supporting their assessment of student assignments.

Figure 9, displayed below, illustrates the output of the auto-evaluator feature. In this scenario, the instructor supplies the system with a key (correct answers) and the students' solutions. The system then generates scores for each question and provides corresponding explanations. While the system's scoring may occasionally be lenient, it still serves as a valuable tool for instructors in making final grading decisions, especially when accompanied by the explanations provided.

To facilitate visual assessment in the grading process, a color-coded system has been employed for question grading. This system highlights questions with a score of less than or equal to two in red, questions with a score of less than or equal to five in yellow, and questions with a score greater than five in green. This color scheme enables instructors to swiftly discern the performance level of each question, thus streamlining the evaluation process.

| Score | Explanation |
|---|---|
| | **Student 1** |
| 2/10 | The student's answer is different from the actual answer because they simply state that they don't know the answer to the question. They don't provide any information or explanation about what a unit is or how it is used. The actual answer provides a clear definition of a unit and explains that it is a standard magnitude used for comparison. It also mentions that there are different units for expressing quantitative values, and the choice of unit depends on the magnitude of the number being expressed. Considering the lack of information in the student's answer, I would grade it a 2 out of 10. |
| 4/10 | The student answer is incorrect because it does not accurately define the relationship between units and dimensions. The student describes units as standard sizes for measuring different things, which is a bit vague and not entirely correct. Additionally, the student defines dimensions as the things that can be measured or counted, which is not a precise definition. In contrast, the actual answer correctly states that a unit is a standard magnitude of a given dimension, against which other magnitudes can be compared. It also defines dimensions as something that can be measured or quantified. The actual answer provides a more accurate and precise explanation of the relationship between units and dimensions. Considering the inaccuracy and lack of precision in the student answer, I would grade it a 4 out of 10. |
| 4/10 | The student's answer is different from the actual answer because it oversimplifies the concept of significant figures and does not accurately convey the message. Firstly, the student's answer states that any digit that isn't just a placeholder zero or doesn't have zeros in front of it is important, while the actual answer specifies that any digit except for zeros used only for the location of the decimal point or those zeros that do not have a nonzero digit on their left are significant digits. This difference is important because it distinguishes between significant and non-significant zeros. Secondly, the student's answer suggests that the more significant figures there are, the more precise the measurement or calculation is. While this is partially true, the actual answer clarifies that the number of significant figures provides an indication of the precision of the measurement. It is not solely determined by the number of significant figures, but also by the instrument used and the limitations of the measurement. Lastly, the student's answer mentions that if an equation doesn't have the right dimensions, it can't be right. Although this statement is true, it is not directly related to the concept of significant figures. The actual answer correctly states that an equation that is not dimensionally homogeneous cannot possibly be valid, but it does not directly connect this concept to significant figures. Overall, I would grade the student's answer a 4 out of 10. While some aspects are correct, there are significant inaccuracies and important details missing that prevent a clear understanding of the concept of significant figures. |
| 6/10 | The student answer is similar to the actual answer in terms of mentioning various units of measurement for length, time, and force. However, the student answer lacks the precision and clarity of the actual answer. Firstly, the student answer begins with a statement that there are "tons of different units out there," which is quite vague and does not provide a specific list of units like the actual answer does. Secondly, the student answer mentions inches, feet, and meters for measuring length, which aligns with the actual answer. However, it fails to mention other units like centimeters or kilometers. Thirdly, the student answer mentions Angstroms and light years for measuring small and large distances in space, which matches the actual answer. However, it does not specify that Angstroms are used for measuring atomic distances. Fourthly, the student answer includes various units for measuring time intervals such as minutes, hours, days, milliseconds, and microseconds, which is similar to the actual answer. However, it omits units like seconds or nanoseconds. Lastly, the student answer mentions Newtons and dynes as units for measuring force, which aligns with the actual answer. However, it does not provide any further explanation or mention any other units for force measurement. Overall, the student answer does touch on various units of measurement but lacks the precision, completeness, and specific examples provided in the actual answer. I would grade the student answer a 6 out of 10. |
| 2/10 | The student answer is different and doesn't convey a similar message to the actual answer. The student simply states that they are unsure of the answer, while the actual answer provides a clear explanation of what a base unit, multiple unit, and derived unit are. Therefore, the student answer is incorrect. Grade: 2/10 |
| 7/10 | The student's answer is almost correct, but there are a few minor inaccuracies. Firstly, the student states that dimensions are "like" things we can measure or quantify, whereas the actual answer states that dimensions "are" something that can be measured or quantified. This difference in wording may seem subtle, but it changes the clarity of the explanation. The actual answer provides a more definitive statement, while the student's answer leaves room for interpretation. Secondly, the student mentions that units are "the specific measurements we use to describe" dimensions. While this is partially true, units are actually the standardized values or labels that we assign to measurements of dimensions. Units provide a specific magnitude or quantity for a given dimension, rather than being a description of the dimension itself. Overall, the student's answer demonstrates a good understanding of the difference between units and dimensions, but the inaccuracies in wording and explanation prevent it from fully conveying the intended message. I would grade the student's answer a 7 out of 10. |
| 1/10 | The student's answer is different from the actual answer because they state that they have no idea what the answer is. They do not provide any information or explanation about the problems an engineer might encounter. Therefore, their answer does not convey a similar message as the actual answer. Grade: 1/10. The student did not provide any relevant information or attempt to answer the question. |
| 7/10 | The student answer is similar to the actual answer in terms of mentioning important stages of the engineering design process such as setting goals and criteria, combining ideas, examining the design, building it, testing it, and evaluating the results. However, the student answer does not use the specific terms "identifying and establishing objectives and criteria" and "synthesis" to describe the initial stages of the process. Additionally, the student answer does not explicitly mention the step of analysis, which is mentioned in the actual answer. Based on these discrepancies, I would grade the student's answer a 7 out of 10. While they touch on the main stages of the engineering design process, they did not accurately use the specific terms provided in the actual answer and missed mentioning the step of analysis. |
| 1/10 | The student answer is different from the actual answer because the student simply states "I'm not sure" without providing any explanation or attempt to answer the question. The actual answer, on the other hand, lists several specific responsibilities that an engineer might have in their workplace. The student's answer does not convey a similar message as the actual answer because it does not provide any information or insight into the topic of engineer responsibilities. Grade: 1/10. The student did not provide an answer or attempt to address the question. |
| 7/10 | The student's answer is similar to the actual answer in terms of the overall structure and content. However, there are a few differences that make the student's answer less accurate. 1. In the student's answer, there is a lack of clarity in some sentences. For example, the sentence "result communication" should be "communication of results" to convey the intended meaning. 2. The student's answer states that prototypes "may be created and tested to meet constraints and criteria," while the actual answer states that prototypes "may be built and tested to meet constraints and criteria." The use of "built" in the actual answer is more accurate and conveys the process of physically constructing prototypes. 3. The student's answer mentions that optimization occurs "after the solution is determined through alternative analysis," while the actual answer states that optimization occurs "after the solution is determined based on the analysis of alternatives." The student's answer is not as clear and does not accurately convey that optimization is based on the analysis of alternatives. 4. The student's answer mentions that a disposal plan "may be necessary before marketing," while the actual answer states that a disposal plan "may be required prior to marketing." The use of "required" in the actual answer is more accurate and emphasizes that a disposal plan is a mandatory step. Overall, I would grade the student's answer a 7 out of 10. While the answer includes most of the key points, there are some inaccuracies and lack of clarity in certain areas. |
| **Total Score: 41/100** | |

Figure 9: Auto-Evaluator Output

### 4.2.2. Automated Homework Detection Mechanism

VirtualTA system incorporates an automatic homework detection feature that caters to the instructor's preferences. When instructors designate certain assignments or homework as off-limits for direct answers, VirtualTA ensures that students seeking assistance related to those specific tasks are guided toward appropriate resources instead. This approach encourages students to engage actively with the course materials and learn the underlying concepts, rather than relying on direct solutions or answers to their homework. By providing guidance and directing students to relevant resources, VirtualTA promotes a deeper understanding of the subject matter, fostering independent learning and critical thinking skills. This feature supports instructors' goals of encouraging academic integrity and facilitating a more comprehensive learning experience for students.

Figure 10, presented below, illustrates the homework detection mechanism. When a question is posed, the system employs a sophisticated algorithm to determine if it resembles a homework or assignment question. If the system detects such a question, it refrains from providing a direct answer but instead guides the students towards appropriate resources where they can seek assistance in answering the question. This mechanism encourages students to engage in independent learning and ensures that they receive the necessary support without compromising the integrity of their academic assignments.

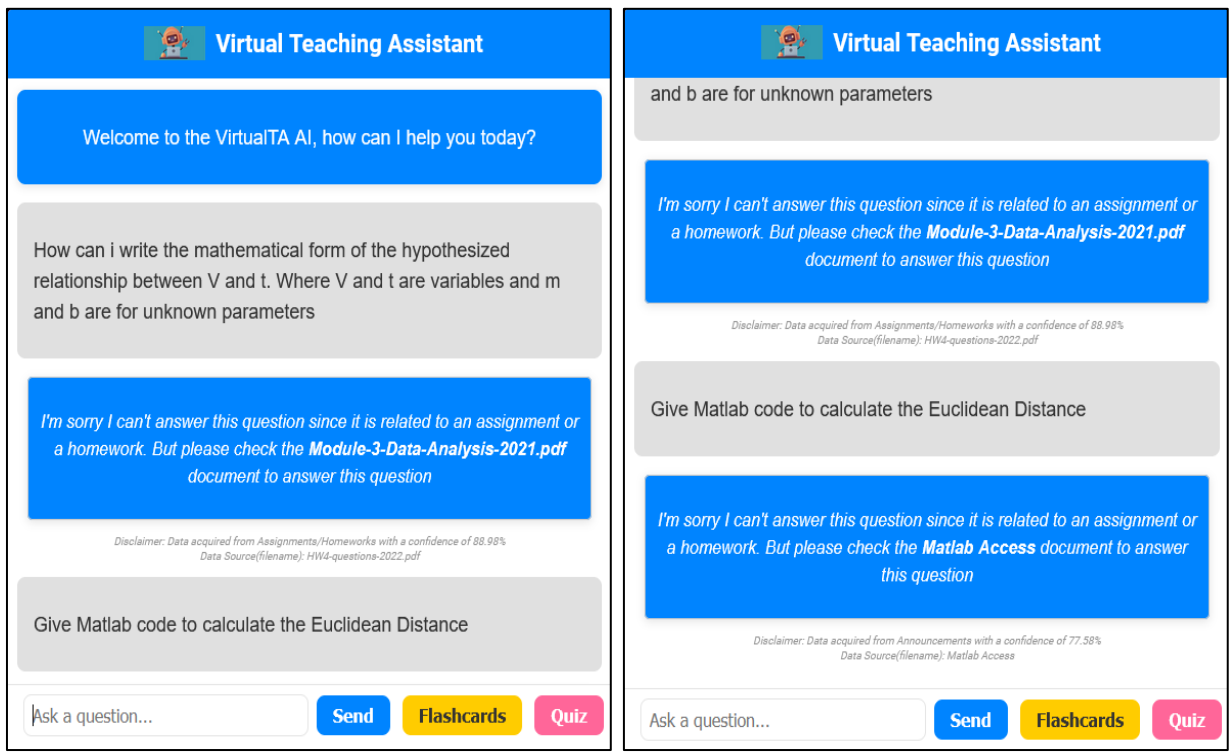

Figure 10: Homework detection mechanism

### 4.2.3. Automated Generation of Diverse Assessment Questions

The system includes a feature that empowers instructors to request VirtualTA to generate questions for exams or quizzes. This functionality serves to streamline the often-laborious task of question generation, providing instructors with a convenient and efficient solution. Instructors have the flexibility to specify the question type, selecting from options such as True/False, Multiple Choice, or Open-Ended questions. By leveraging this feature, instructors can save valuable time and effort, allowing them to focus on other aspects of course preparation and instruction. This capability offered by VirtualTA aims to enhance the overall experience for instructors, facilitating the creation of diverse and relevant assessment materials.

## 5. Discussions

VirtualTA system offers a more comprehensive and adaptable approach to educational support compared to existing educational chatbot systems. Its integration of various functionalities, support for different question types, context-awareness, and emphasis on academic integrity and learning analytics contribute to a more sophisticated and effective educational support system. In comparison to existing research and applications in the field of educational chatbots, VirtualTA system introduces novel features and addresses specific challenges in higher education, proving it to be a valuable contribution to the field of educational technology and chatbot development.

Firstly, VirtualTA system goes beyond traditional chatbot functionalities by incorporating features such as flashcards, quizzes, automated homework evaluation, coding sandbox, and summary generation. These additional functionalities provide a comprehensive learning support ecosystem that goes beyond basic question-answering capabilities. Secondly, VirtualTA system emphasizes the importance of academic integrity and learning analytics. By providing automated homework evaluation and incorporating measures to prevent cheating, the system ensures fair assessment and promotes ethical academic practices. The utilization of learning analytics enables instructors to gain insights into student performance and engagement, facilitating data-driven decision-making. Furthermore, VirtualTA system aims to integrate seamlessly with existing Learning Management Systems (LMS), such as Canvas, to enhance accessibility and user experience. This integration potential sets it apart from standalone chatbot systems and allows for a more integrated and streamlined educational environment.

### 5.1. Limitations and Challenges

While VirtualTA system demonstrates great potential, it is important to acknowledge the challenges and limitations encountered during its development. Throughout the development of VirtualTA system, we encountered several challenges and limitations that shaped the implementation. One major challenge we faced was the handling of PDF files, which often contain unstructured data. Extracting structured information from PDFs proved to be a complex task, especially when dealing with scanned copies that require Optical Character Recognition (OCR) to parse the content accurately. While we did not implement OCR functionality at the time of writing this paper, it remains a limitation that can be addressed in future iterations of the system.

Another challenge we encountered was related to the integration of Learning Management Systems (LMS). LMS platforms typically lack standardized methods for requesting data in a desired format. As a result, we had to devise workarounds to extract and process the necessary information from the LMS. This required careful development of a custom LMS library to ensure compatibility and efficient data retrieval. Additionally, integrating the Whisper ASR system posed challenges due to the limitations of the API. The API imposes a constraint of 25MB on the data size (Brockman et al., 2023), while many class recordings, including video files (MP4), exceed this limit. To overcome this limitation, the videos were partitioned into smaller chunks or compressed to reduce the file size, enabling its utilization within the Whisper API.

Furthermore, the frequent updates and advancements in the underlying models posed another challenge. As the models evolved, we needed to upgrade the APIs and adapt the system to leverage the latest technological improvements. Staying abreast of the newer developments in the field required continuous effort to ensure VirtualTA system remained up-to-date and aligned with the state-of-the-art techniques. These challenges and limitations underscore the iterative nature of the system development, where ongoing improvements and future enhancements can address these areas and further enhance the system's capabilities.

### 5.2. Opportunities and Future Directions

The research findings and development of VirtualTA system open numerous opportunities and future directions for further improvements. By addressing these opportunities and future directions, VirtualTA system can further revolutionize the role of AI in higher education, enhancing student learning experiences, and paving the way for the next generation of educational technology.

a) <u>Enhanced Natural Language Understanding:</u> Invest in research and development to improve the system's natural language understanding capabilities by exploring advanced natural language processing techniques, such as semantic parsing, entity recognition, and sentiment analysis.
b) <u>Personalization and Adaptive Learning:</u> Develop adaptive learning algorithms to personalize VirtualTA system, addressing the unique needs and learning styles of each student, fostering engagement, and contributing to more effective learning outcomes.
c) <u>Multimodal Learning Support:</u> Integrate multimedia resources, such as video lectures, interactive simulations, and visual aids, to provide comprehensive and diverse learning support for various learning styles and preferences.
d) <u>User Feedback and Evaluation:</u> Conduct rigorous user feedback and evaluation studies to gather insights into VirtualTA system's effectiveness and usability. Feedback from students, instructors, and educational stakeholders will help identify areas of improvement and validate the system's impact on student learning outcomes.
e) <u>Integration with Multiple LMSs:</u> Investigate the feasibility of integrating the AIIA with a broader range of LMS, ensuring compatibility with various institutions and expanding its reach.
f) <u>Real-time Video Interaction:</u> Implement real-time video interaction features, enabling students to virtually attend lectures, ask questions, and receive immediate feedback from the AI assistant or instructors.
g) <u>Instructor-Assistant Collaboration:</u> Enhance VirtualTA system to include features that foster collaboration between instructors and the AI assistant, allowing them to share content, coordinate responses, and provide combined support to students.

h) <u>Gamification and Engagement:</u> Integrate gamification elements within VirtualTA system to motivate students, enhance engagement, and create a more enjoyable learning experience.
i) <u>Longitudinal Studies:</u> Conduct long-term studies to assess the impact of VirtualTA system on student performance, retention, and overall academic outcomes.
j) <u>Ethical Considerations and Privacy:</u> Investigate the ethical implications of using AI in education, addressing concerns related to data privacy, algorithmic bias, and the potential impact on the human role in education.

## 6. Conclusions

This research has presented the design, implementation, and evaluation of an Artificial Intelligence-Enabled Intelligent Assistant (AIIA) for personalized and adaptive learning in higher education. Through the integration of advanced AI technologies and natural language processing techniques, VirtualTA system aims to enhance learning outcomes and promote student engagement, while addressing the diverse needs of learners in qualitative disciplines. The system's capabilities span various functionalities, including responsive question-answering, flashcard integration, automated assessment, embedded coding sandbox, summarization of course content, and context-aware conversation. Additionally, the system offers instructor-focused enhancements, such as auto-evaluation for assignment grading, homework detection mechanisms, and automated question generation.

By providing a comprehensive suite of tools and resources, VirtualTA system has the potential to revolutionize the role of AI in higher education. However, it is crucial to acknowledge the challenges and limitations encountered during the development process, which can be addressed in future iterations. The opportunities and directions outlined in this paper provide a roadmap for further advancements in VirtualTA system and the broader field of AI-enabled educational technology.

In conclusion, VirtualTA system represents a significant contribution to the ongoing efforts to integrate AI and natural language processing into educational contexts. By fostering self-regulated learning, promoting student-faculty communication, and expanding access to learning resources, the AIIA framework aims to enhance the effectiveness of learning support and shape the future trajectory of higher education. As we continue to refine the system and explore new avenues of research and development, we move closer to realizing the full potential of AI-enabled educational technology in transforming the higher education landscape, empowering learners, and nurturing the next generation of professionals.

**Funding**

Funding for this project was provided by the National Oceanic & Atmospheric Administration (NOAA), awarded to the Cooperative Institute for Research to Operations in Hydrology (CIROH) through the NOAA Cooperative Agreement with The University of Alabama (NA22NWS4320003) and National Science Foundation (#2230710).

**Availability of Data and Materials**

All data that is produced and analyzed in the manuscript is readily available and presented in the manuscript.